\documentclass[letterpaper, 10 pt, conference]{ieeeconf}

\IEEEoverridecommandlockouts 
\usepackage{cite}
\usepackage{graphics} % for pdf, bitmapped graphics files
\usepackage{epsfig} % for postscript graphics files
\usepackage{amsmath} % assumes amsmath package installed

\usepackage{subcaption}
\usepackage{float}

\usepackage{multirow}
\usepackage{booktabs}

\usepackage{tikz}
\usepackage{amssymb}

\usepackage{hyperref}

\hypersetup{
    colorlinks=true,
    linkcolor=blue,
    filecolor=magenta,
    urlcolor=cyan,
    citecolor=green,
}
\hypersetup{
    hidelinks,
}

\definecolor{burgundy}{rgb}{0.5, 0.0, 0.13}

\usepackage{fancyhdr}

\fancypagestyle{firstpage}{%
  \fancyhead{}
  
  \fancyfoot[C]{\textit{This work has been submitted to the IEEE for possible publication.\\
Copyright may be transferred without notice, after which this version may no longer be accessible.}}
}

\newcommand{\cg}[2]{
  \begin{subfigure}[t]{0.147\textwidth}
    \centering
    \includegraphics[width=\textwidth]{content/images/cg/r#1_1.png}% 
    \vspace{-5pt}
    \caption*{\textit{``#2''}}
  \end{subfigure}
  \hspace{-8pt}
  \begin{subfigure}[t]{0.147\textwidth}
    \centering 
    \includegraphics[width=\textwidth]{content/images/cg/r#1_3.png}% 
    \vspace{-5pt}
    \caption*{ConceptGraph \cite{conceptgraph}}
  \end{subfigure}
  \hspace{-8pt}
  \begin{subfigure}[t]{0.147\textwidth}
    \centering
    \includegraphics[width=\textwidth]{content/images/cg/r#1_2.png}% 
    \vspace{-5pt}
    \caption*{OpenSU3D (Ours)}
  \end{subfigure}%
  }

\newcommand{\od}[2]{
  \begin{subfigure}[t]{0.147\textwidth}
    \centering
    \includegraphics[width=\textwidth]{content/images/o3d/r#1_1.png}% 
    \vspace{-5pt}
    \caption*{\textit{``#2''}}
  \end{subfigure}
  \hspace{-8pt}
  \begin{subfigure}[t]{0.147\textwidth}
    \centering 
    \includegraphics[width=\textwidth]{content/images/o3d/r#1_3.png}% 
    \vspace{-5pt}
    \caption*{OpenMask3D \cite{openmask}}
  \end{subfigure}
  \hspace{-8pt}
  \begin{subfigure}[t]{0.147\textwidth}
    \centering
    \includegraphics[width=\textwidth]{content/images/o3d/r#1_2.png}% 
    \vspace{-5pt}
    \caption*{OpenSU3D (Ours)}
  \end{subfigure}%
  }

\newcommand{\schmshort}[2]{
  \caption*{\textbf{Query: }\textit{``#2''}}
  \begin{subfigure}[t]{0.121\textwidth}
    \centering 
    \includegraphics[width=\textwidth]{content/images/schm2/r#1_2.png}% 
    \vspace{-5pt}
    \caption*{Scheme 1}
  \end{subfigure}
  \hspace{-8pt}
  \begin{subfigure}[t]{0.121\textwidth}
    \centering
    \includegraphics[width=\textwidth]{content/images/schm2/r#1_3.png}% 
    \vspace{-5pt}
    \caption*{Scheme 2}
  \end{subfigure}%
  \hspace{-8pt}
  \begin{subfigure}[t]{0.121\textwidth}
    \centering
    \includegraphics[width=\textwidth]{content/images/schm2/r#1_4.png}% 
    \vspace{-5pt}
    \caption*{Scheme 3}
  \end{subfigure}%
  \hspace{-8pt}
  \begin{subfigure}[t]{0.121\textwidth}
    \centering
    \includegraphics[width=\textwidth]{content/images/schm2/r#1_5.png}% 
    \vspace{-5pt}
    \caption*{Scheme 4}
  \end{subfigure}%
  }

\usepackage{listings}

\usepackage{fancyvrb,fvextra}
\usepackage{spverbatim}
\usepackage{ragged2e}
\usepackage{cite}

\title{\LARGE \bf
OpenSU3D: Open World 3D Scene Understanding using \\ Foundation Models}
\author{Rafay Mohiuddin$^{1*}$, Sai Manoj Prakhya$^{2}$, Fiona Collins$^{1}$, Ziyuan Liu$^{2}$ and Andr\'e Borrmann$^{1}$% <-this % stops a space
\thanks{$^{*}$ Corresponding author \texttt{rafay.mohiuddin@tum.de}}% <-this % stops a space
\thanks{$^{1}$ Chair of Computational Modeling \& Simulation, Technical University of Munich, 80333 Arcisstraße 21, Germany}%
\thanks{$^{2}$ Intelligent Cloud Technologies Lab, Huawei Munich Research Center, 80992 Riesstraße 25, Germany}%
}

\begin{document}

\maketitle
\thispagestyle{empty}
\pagestyle{empty}

\thispagestyle{firstpage}

%%%%%%%%%%%%%%%%%%%%%%%%%%%%%%%%%%%%%%%%%%%%%%%%%%%%%%%%%%%%%%%%%%%%%%%%%%%%%%%%

\vspace{-0.3cm} 
\begin{figure*}[h]
  \centering

    \includegraphics[width=0.77\textwidth,keepaspectratio]{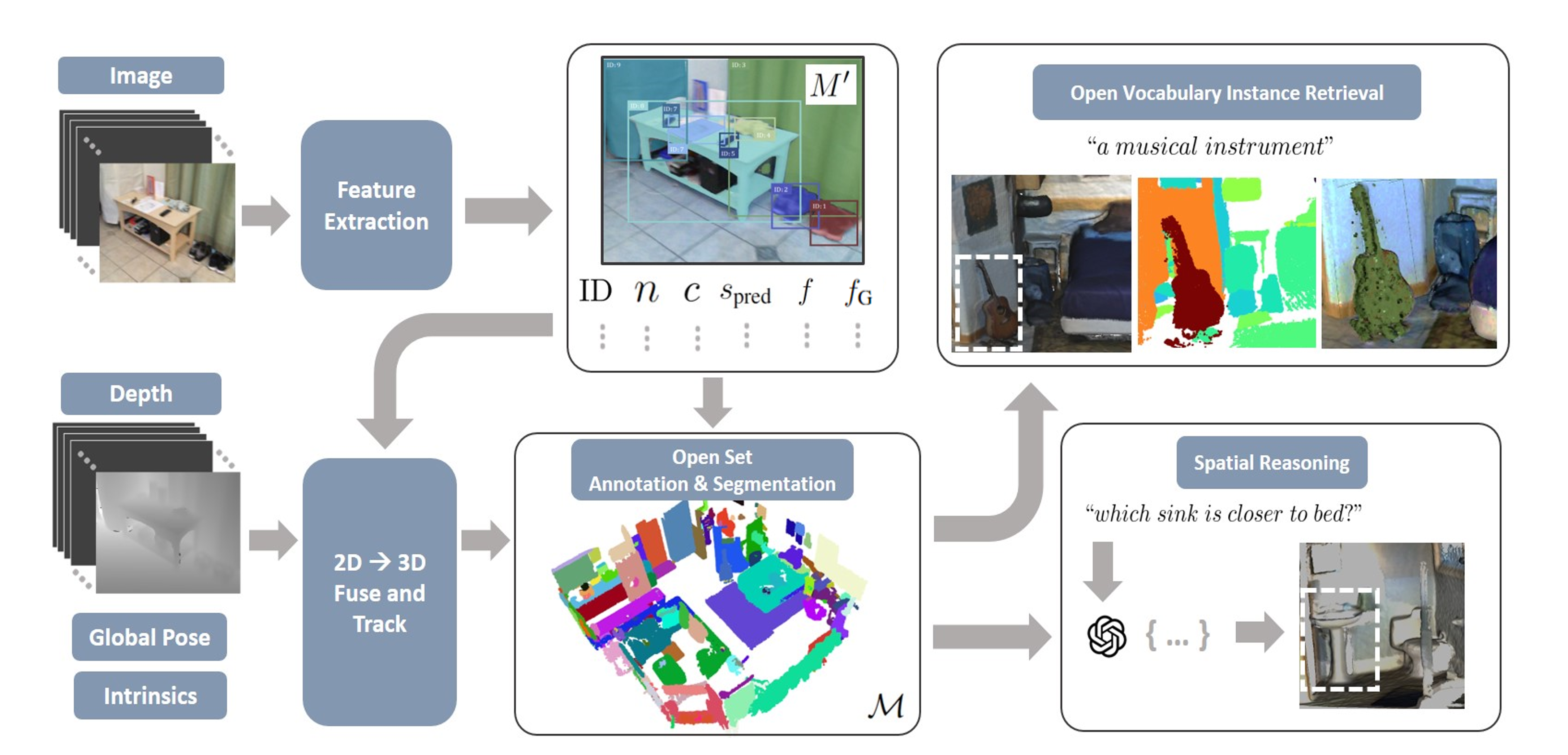}

\caption{\textbf{Open World 3D Scene Understanding Pipeline.} Our method takes a sequence of RGB-D images and constructs a 3D scene representation for open vocabulary instance retrieval, open set annotation, segmentation, and spatial reasoning.}

\vspace{-0.5cm} % Adjust this value to reduce the gap

  \label{fig:flow}
\end{figure*}

\begin{abstract}

In this paper, we present a novel, scalable approach for constructing open set, instance-level 3D scene representations, advancing open world understanding of 3D environments. Existing methods require pre-constructed 3D scenes and face scalability issues due to per-point feature representation, additionally struggle with contextual queries. Our method overcomes these limitations by incrementally building instance-level 3D scene representations using 2D foundation models, and efficiently aggregating instance-level details such as masks, feature vectors, names, and captions. We introduce fusion schemes for feature vectors to enhance their contextual knowledge and performance on complex queries. Additionally, we explore large language models for robust automatic annotation and spatial reasoning tasks. We evaluate our proposed approach on multiple scenes from ScanNet \cite{scannet} and Replica \cite{replica} datasets demonstrating zero-shot generalization capabilities, exceeding current state-of-the-art methods in open world 3D scene understanding. Project page: \href{https://opensu3d.github.io/}{\texttt{\textcolor{purple}{https://opensu3d.github.io/}}}
\end{abstract}

%\newline 

%\begin{keywords}
%3D Scene understanding, Instance Segmentation, Automatic Labeling, Spatial Reasoning
%\end{keywords}

%%%%%%%%%%%%%%%%%%%%%%%%%%%%%%%%%%%%%%%%%%%%%%%%%%%%%%%%%%%%%%%%%%%%%%%%%%%%%%%%
%\section{INTRODUCTION}
\section{INTRODUCTION}
\label{ch:intro}

% target -> 85-100 lines
%\subsection{Motivation}
% 2.1 Importance and General Discussion
    % 2D to 3D information transfer
    % closed set short comming
    % application for robotics
    % application for digital twin

%\subsubsection{Motivation}

Recent advancements in AI, particularly in open-set object detection and contextual understanding of 2D images, owe much to pre-trained foundation models like CLIP \cite{clip}, SAM \cite{sam}, and the integration of vision with language models \cite{llava, gpt4}. However, extending these breakthroughs to 3D scenes remains a challenge. While innovative, current 3D methods \cite{3dclr,openscene,3dllm,pla,conceptfusion} have not yet achieved the performance levels seen in 2D. Addressing this gap is critical for robotics, as it could transform how robots perceive, interact, and reason within the three-dimensional world.

Recent efforts \cite{3dclr,openscene,3dllm,pla,conceptfusion} have made strides in adapting 2D foundation models for open-world 3D scene understanding. However, they face key limitations. Many are designed for batch-processing or non-incremental tasks, requiring complete 3D scene data upfront—unrealistic in many real-world robotics applications. These approaches primarily derive per point 3D feature vectors from 2D models like CLIP \cite{clip, clipseg}, but lack a generalized approach to extrapolate 2D information from other foundational models. Moreover, generating dense, per-point feature representations raises memory and scalability challenges while complicating the task of isolating distinct entities in a scene, critical for robotic operations. Most notably, the existing methods seem to work effectively for simple queries but lack the depth and contextual understanding required for more complex spatial queries and reasoning tasks.%These methods, while effective for simple queries, struggle with more complex spatial query and reasoning tasks.

\subsection{Overview of Contributions}
\vspace{-0.1cm}
We present a novel approach for constructing open-set 3D scene representation addressing \textbf{open vocabulary instance recall, segmentation, annotation, and spatial reasoning}. Our method leverages 2D foundation models, specifically GroundedSAM~\cite{gsam} and GPT-4V~\cite{gpt4}, to extract instance-level information from RGB images.
For each instance, CLIP~\cite{clip} feature vectors are extracted and fused at multiple scales.
Unique IDs are assigned to each instance in an image; corresponding 2D masks are back-projected using depth and pose data to construct a segmented 3D scene. Our approach tracks association between 2D and 3D masks by assessing overlapping region in 3D space, and updates 3D mask using corresponding meta-information from 2D images, enabling efficient, scalable, and incremental 3D scene construction. Additionally,  feature fusion schemes incorporate local context, aiding in distinguishing instances within the same class for relational queries. 
%\textbf{Key Contributions:}
This study brings the following key contributions to the field of 3D scene understanding:
\newline\textbf{1.}  We introduce an incremental and scalable approach for open-set 3D scene understanding and instance segmentation, integrating instance-level information from 2D foundational models into a unified 3D representation.
\newline\textbf{2.}  We develop an innovative feature fusion formulation encompassing contextual information for improved instance identification through contextual queries.
\newline\textbf{3.}  We explore the use of large language models in conjunction with our 3D scene representations for robust automatic annotation and complex spatial reasoning tasks
    
%\subsubsection{Proposed Approach}
 %Evaluations on 

\vspace{-0.2cm}

%\section{RELATED WORK}

\subsection{Literature Review}
\label{sec:rel_work}
\vspace{-0.1cm}
\subsubsection{Foundation \& Large Language Models} Foundation models, trained on large-scale data, have transformed AI by achieving exceptional performance across varied tasks. CLIP ~\cite{clip}, and \cite{blip, blip2} integrate visual and textual information into a unified representation, enabling multimodal tasks such as image captioning, visual question answering, and cross-modal retrieval. In segmentation, ~\cite{Lseg, ovseg} provide promptable, open-vocabulary capabilities. Grounding approaches, ~\cite{gsam, seem}, build on SAM~\cite{sam} contextualize outputs by incorporating semantic and situational information, improving interpretative accuracy and relevance in tasks like image segmentation, captioning and object detection. Language models~\cite{gpt3,llama} excel in natural language tasks, and their integration with vision~\cite{llava2,gpt4} further advances open-world comprehension. 
%This work explores an efficient and generalizable framework for linking 2D images with 3D spaces, transferring the capabilities of 2D foundation models and large language models.
This work explores generalizable approach for extracting and linking information between 2D images and 3D spaces, leveraging the capabilities of foundation and large language models.

\subsubsection{3D Scene Segmentation} 
Semantic segmentation remains a critical challenge in 3D robotics perception. While approaches like \cite{voxblox, kimera, hydra, sgfusion} creates 3D metric semantic map, they are limited by closed-set paradigms. Recent works~\cite{isgfusion,sam3d} employ simplified approach, identifying 3D instances through overlapping 3D points from semantically segmented RGB images. ~\cite{sam3d} produces non-incrementally open-set fine-grained 3D masks, while~\cite{isgfusion} creates an incremental, closed-set sparse 3D semantic map with fixed per-update computation. We propose an incremental method leveraging SAM's~\cite{sam} 2D masks and region overlap techniques to generate fine-grained 3D instance masks with constant per-update computation. Additionally, our approach efficiently tracks 2D-3D mask correspondence, facilitating effective instance-level information transfer from 2D foundation models to 3D scenes.

\vspace{0.01cm}

\subsubsection{Open Vocab. 3D Scene Understanding}
Recent advancements in 3D scene understanding have leveraged 2D vision-language models to enable open-vocabulary interaction. While early works ~\cite{openscene, conceptfusion, pla, lerf} demonstrated potential, they faced computational and scalability challenges due to dense feature representations. Subsequent approaches, ~\cite{openmask, openins3d}, adopted instance-centric methods to address these issues. However, their non-incremental nature and reliance on pre-constructed 3D scene limits applicability in robotics. Additionally, ~\cite{conceptfusion, openmask} explores feature engineering to improve feature representation, fusing CLIP~\cite{clip} vectors from object-centric and larger image sections but only shows performance on simple queries. Despite progress, challenges persist in computation, scalability, incremental processing, and contextual query handling, necessitating more efficient and adaptable solutions.

\subsubsection{3D Spatial Reasoning}
Global 3D spatial reasoning remains a challenge in open-world scene understanding. While conventional scene graph approaches~\cite{hydra,sgfusion} show limitations, recent works~\cite{3dclr,3dllm,charwithnerf} leverage LLMs for 3D reasoning. However, even advanced models like GPT-4V~\cite{gpt4} struggle with accurate 2D spatial reasoning~\cite{som}. Adapting the Set of Mark Prompting technique~\cite{som} for 3D environments, we employ LLMs for 3D spatial reasoning over constructed scenes through strategic prompting.

\subsubsection{Recent Related Works}
%Recent work N2F2~\cite{n2f2}, LangSplat~\cite{langsplat}, Segment3D~\cite{segment3d}, OpenIns3D~\cite{openins3d}, OVSG~\cite{ovsg}, ConceptGraph~\cite{conceptgraph} and Cilo~\cite{cilo},
Recent works~\cite{n2f2,langsplat,segment3d,openins3d,ovsg,conceptgraph,cilo} 
also addresses 3D scene understanding. Unlike ~\cite{ovsg} and ~\cite{conceptgraph},  our method ensures semantic consistency in 3D masks through geometric overlap of 3D points rather than CLIP feature vectors, achieving better performance as shown in experiments. While ~\cite{segment3d, openins3d}, are non-incremental, ~\cite{n2f2,langsplat} struggle with contextual queries. ~\cite{ovsg,conceptgraph,cilo} employ scene graphs and LLMs for spatial queries; in contrast, we explore in-context learning, leveraging large context length of LLMs and tailored prompting for complex spatial reasoning tasks.

\vspace{-0.1cm}

%\section{METHOD}
\section{METHOD}
\label{sec:the_method}
Our approach processes RGB-D image sequences with corresponding poses to generate an open-set 3D scene representation, supporting open-world tasks such as open-vocabulary object retrieval, 3D segmentation, annotation, and spatial reasoning. As shown in Fig.\ref{fig:flow}, the pipeline consists of two key modules:
\newline\textbf{1. Per-Image Feature Extraction:} Extracts instance-level masks, embeddings, and meta-information from each image and assigns a unique \text{ID} to each instance for precise tracking.
\newline\textbf{2. 2D to 3D Fusion and Tracking:} Creates a 3D semantic map from per-image 2D masks and associated 2D meta-information into 3D space by tracking the corresponding IDs.

\subsection{Per-Image Feature Extraction}
\label{sec:method_feat_extract} %$\mathcal{I'} = \{I_0, I_{s}, I_{2s}, \ldots , I_n\}$
The feature extraction process, begins with a sequence of RGB images, $\mathcal{I} = \{I_0, I_{1}, I_{2}, \ldots , I_n\}$. A subset $\mathcal{I'}$ is sampled with a stride $s$ to minimize computational redundancy while ensuring a reasonable overlap. For each image $I' \in \mathcal{I'}$, groundedSAM~\cite{gsam} is used to obtain 2D masks $M$, bounding boxes $BB$, and prediction scores $S_{\text{pred}}$. Crops from image, based on $bb \in BB$, are passed to GPT-4V~\cite{gpt4} with prompt \textit{``Give name and description to the object. Output e.g. `name':`description'"} to get names (labels) $N$ and detailed captions $C$ describing the objects. Each instance is assigned a unique ID, and masks $M$ are updated to $M'$ with these IDs as per-pixel labels, including a border of $px$ pixels around each mask to delineate entities. Feature vectors are extracted using the CLIP encoder in two stages:
\begin{enumerate}
  \item A global feature vector $f_{\text{G}}$ is extracted from full image.
  \item Instance-specific feature vectors $F=\{f_{\text{MS}}\}$  are created by first cropping the image at multiple scales based on scaling ratios $S_r = \{s_{r}\}_{k}$ and then fusing per-crop feature vectors $\{f\}_k$  with a multiscale feature fusion scheme as discussed in Sec. \ref{sec:feature_fusion}.
\end{enumerate}

The updated masks $M'$, and instance-level metadata including IDs, names $n \in N$, captions $c \in C$, prediction scores $s_{\text{pred}} \in S_{\text{pred}}$, fused feature vectors $f_{\text{MS}} \in F$, and global feature vector $f_{\text{G}}$, are stored for each image in $\mathcal{I'}$.%\textbf{ Please note that I have changed $f$ to $f_{i}$ in the above paragraph. If there are references to $f$, please change them accordingly for consistency}

\vspace{-0.3cm} 
\begin{figure}[h]
  \centering
    \centering % Centers the image
    \includegraphics[width=0.35\textwidth,keepaspectratio]{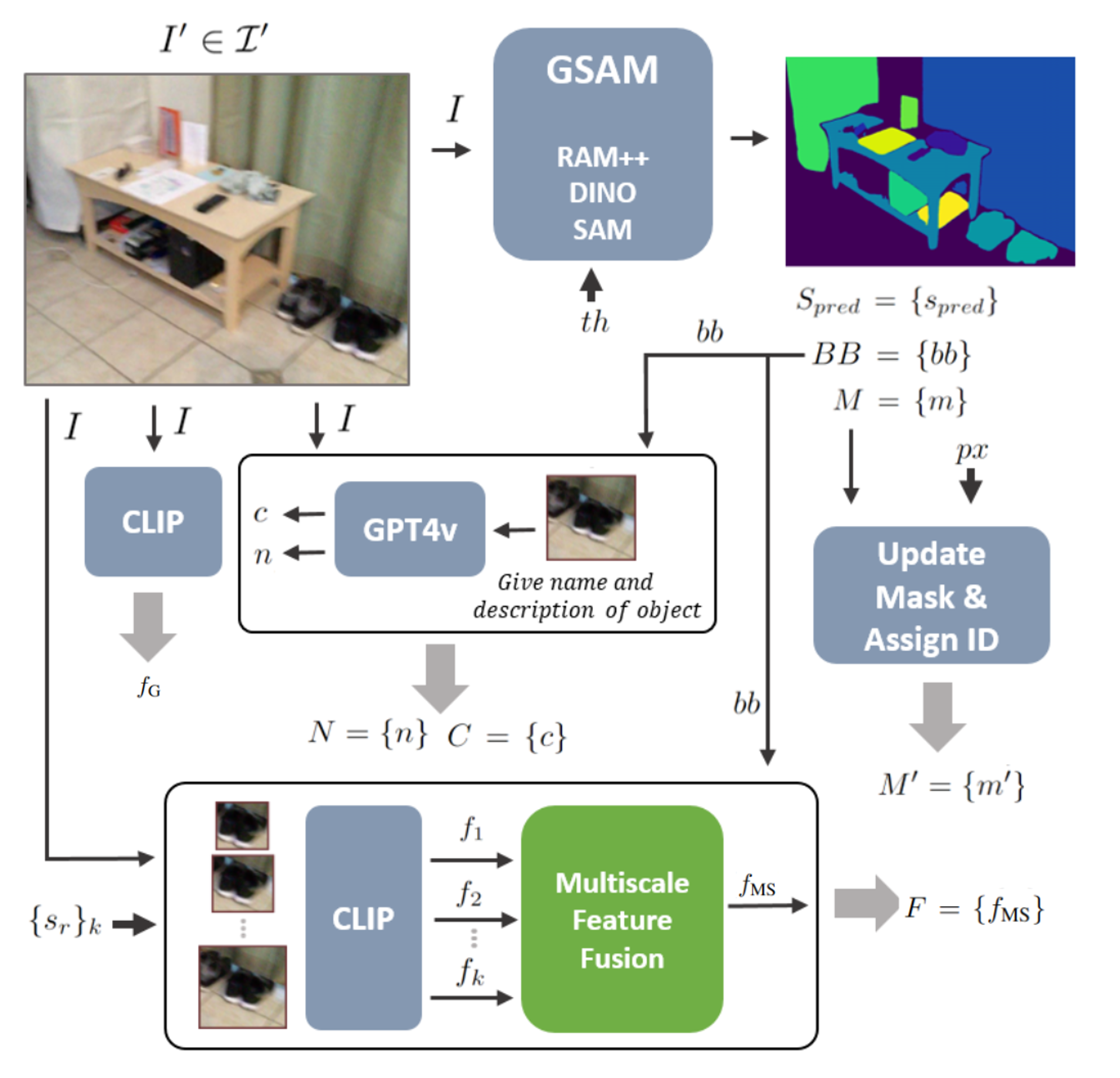}
    %fe
    %%% Explained
    %\caption{\textbf{Feature extraction:} Using Grounded-SAM ~\cite{gsam}, we extract masks and labels for each image. The labels are further refined and mask description is obtained using GPT-4V. Lastly, a global per-image CLIP feature and fused multi-scale per-mask CLIP features are extracted and stored.}
  \caption{\textbf{Feature Extraction Module.} For each instance in an image assigns a unique ID and extracts name, bounding box, detailed caption, prediction score, and CLIP~\cite{clip} features.}

  %\textbf{Please give more description explaning each step in pipeline that can guide readers clearly}}
% Adjust this value to reduce the gap
  \label{fig:feat_extract}
\end{figure}
\vspace{-0.53cm} 

\subsection{2D to 3D Fusion \& Tracking}
\label{sec:method_2dto3d track}

We initiate the fusion and tracking module by initializing an empty 3D point cloud for the complete 3D scene, represented as $\mathcal{P}_{\text{scene}} \in \mathbb{R}^{x,y,z,\text{ID}}$, and a global hash table $\mathcal{Q}$ for tracking the unique IDs, defined as: $\mathcal{Q} : \mathcal{Q} \mapsto \left\{ \text{ID} \in \texttt{uniq}(\text{ID} \in \mathcal{P}_{\text{scene}}) : \{\text{ID} \in \{M'\}\} \right\}$

For the image $I'$, associated elements including depth maps $D$, global poses $T$, updated masks $M'$, and camera intrinsic $K$ are retrieved. 
%%%\textbf{FYI - backprojection is used to project 3D to images, so it should be like this.}
Each pixel $(u,v) \in I'$ is back-projected into 3D space, using depth data and is assigned a semantic label corresponding to the mask $M'$, resulting in a 3D point cloud for a single image $\mathcal{P}_{\text{frame}}$.
\begin{equation}
    \mathcal{P}_{\text{frame}} = \left\{ T \left( D(u, v) \cdot K^{-1} \begin{pmatrix} u \\ v \\ 1 \end{pmatrix} \right), M'(u, v) \right\}
    \label{eq:backproject}
\end{equation}

%Index pairs \(\{(\mathbf{i}_{\text{frame}}, \mathbf{i}_{\text{scene}})\}\) are used to identify corresponding points between \(\mathcal{P}_{\text{frame}}\) and \(\mathcal{P}_{\text{scene}}\). 
%Using the bounds of \(\mathcal{P}_{\text{frame}}\), we sample \(\mathcal{P}'_{\text{scene}}\) from \(\mathcal{P}_{\text{scene}}\), containing only points within the bounds of \(\mathcal{P}_{\text{frame}}\), reducing the search space for efficient computation.

%%\textbf{Index pairs $\{(\mathbf{i}_{\text{frame}}, \mathbf{i}_{\text{scene}})\}$ are determined to identify corresponding points between $\mathcal{P}_{\text{frame}}$ and $\mathcal{P}_{\text{scene}}$. Using bounds of $\mathcal{P}_{\text{frame}}$, $\mathcal{P}'_{\text{scene}}$ is sampled from  $\mathcal{P}_{\text{scene}}$, containing only points within the bounds of $\mathcal{P}_{\text{frame}}$. - These two sentences need to be re-written. Second sentence is making sense. However, the first sentence is confusing. Please re-write either first or both sentences more clearly}
%A $KDTree$ search is performed, utilizing an Euclidean distance function $d(\cdot,\cdot)$ to matches points $\mathbf{p}$ in $\mathcal{P}_{\text{frame}}$ with points $\mathbf{q}$ of $\mathcal{P}_{\text{scene}}$ in $\mathcal{P'}_{\text{scene}}$ to get corresponding index pairs $\{(\mathbf{i}_{\text{frame}}, \mathbf{i}_{\text{scene}})\}$, where $d(\mathbf{p}, \mathbf{q}) < \epsilon$. This search strategy restricts the computation complexity of our method to  $\mathcal{O}(1)$.
Using the bounds of \(\mathcal{P}_{\text{frame}}\), we sample \(\mathcal{P}'_{\text{scene}}\) from \(\mathcal{P}_{\text{scene}}\), containing only points within the bounds of \(\mathcal{P}_{\text{frame}}\). A $KDTree$ search is performed, utilizing Euclidean distance function, $d(\cdot,\cdot)$ to match points $\mathbf{p}$ $\in$ $\mathcal{P}_{\text{frame}}$ with points $\mathbf{q}$ $\in$ $\mathcal{P'}_{\text{scene}}$. 
%%\textbf{If $d(\mathbf{p}, \mathbf{q}) < \epsilon$, group indices corresponding to $\mathbf{p}$ $\in$ $\mathcal{P}_{\text{frame}}$ with $\mathbf{q}$ $\in$ $\mathcal{P}_{\text{scene}}$ to get respective index pairs $\{(\mathbf{i}_{\text{frame}}, \mathbf{i}_{\text{scene}})\}$ for all overlapping points. - please rewrite clearly} 
If \(d(\mathbf{p}, \mathbf{q}) < \epsilon\), we group the indices corresponding to \(\mathbf{p} \in \mathcal{P}_{\text{frame}}\) with \(\mathbf{q} \in \mathcal{P}_{\text{scene}}\) to obtain the respective index pairs \(\{(\mathbf{i}_{\text{frame}}, \mathbf{i}_{\text{scene}})\}\) for all overlapping points.
This search strategy limits the search space of $KDTree$ to only overlapping region, thereby requiring a constant computation (search space) per update.
%, as opposed to SAM3D \cite{sam3d} that does KDTree search over entire scene. %\textbf{(Are you sure that SAM3D requires logarithmic increase or exponential increase ?)} %constraining the computational complexity of our method to $\mathcal{O}(1)$.

%By subsampling $\mathcal{P}'_{\text{scene}}$ to contain only points within the bounds of $\mathcal{P}_{\text{frame}}$, a $KDTree$ search, utilizing an Euclidean distance function $d(\cdot,\cdot)$, matches indices corresponding points $\mathbf{p}$ in $\mathcal{P}_{\text{frame}}$ with points $\mathbf{q}$ in $\mathcal{P}_{\text{scene}}$ where $d(\mathbf{p}, \mathbf{q} \in \mathcal{P'}_{\text{frame}}) < \epsilon$ (Algo. \ref{algo:index_pairs}).

To track and update matched IDs, similar to SAM3D's \cite{sam3d} approach, we begin by obtaining a list of unique IDs \(\{\text{ID}_{f}\}\) for each 3D segment and a corresponding list denoting the total point count \(\{c_{\mathcal{P}_f}\}\) of each 3D segment in \(\mathcal{P}_{\text{frame}}\).

%%To track and update matched IDs, similar to SAM3D's \cite{sam3d} approach, we begin by obtaining a list of unique IDs $\{\text{ID}_{f}\}$ for each segment and a corresponding list denoting the \textbf{total point count - u need to use a better word that sounds meaningful }$\{c_{\mathcal{P}_f}\}$ of each segment of $\mathcal{P}_{\text{frame}}$.
%This is achieved using the function \texttt{getUniqCount}, which operates on the set of points $\mathcal{P}_{\text{frame}}$ in the frame:
%\begin{equation}
%\{\text{ID}_{f}\}, \{c_{\mathcal{P}_f}\} \gets \texttt{getUniqCount}(\text{ID} \in \mathcal{P}_{\text{frame}})
%\label{eq:get_count_f}
%\end{equation}

%$c_{\mathcal{P}_f} \in \{c_{\mathcal{P}_{f}}\}$

For each 3D segment in \(\mathcal{P}_{\text{frame}}\) with \(c_{\mathcal{P}_f} \in \{c_{\mathcal{P}_{f}}\}\), we utilize the index pairs \(\{(\mathbf{i}_{\text{frame}}, \mathbf{i}_{\text{scene}})\}\) to obtain the set of points from \(\mathcal{P}_{\text{scene}}\) that overlaps with \(\mathcal{P}_{\text{frame}}\). From these points, we derive a list of unique segment IDs \(\{\text{ID}_{s}\}\) and their corresponding total point counts \(\{c_{\mathcal{P}_s}\}\).

\vspace{-0.2cm}
\begin{equation}
\text{OverlapRatio} = \frac{\max(\{c_{\mathcal{P}_s}\})}{\min(c_{\mathcal{P}_f}, \max(\{c_{\mathcal{P}_s}\}))}
\label{eq:overlap_ratio}
\end{equation}
\vspace{-0.2cm}

If the overlap ratio satisfies a pre-defined threshold, i.e., $\text{OverlapRatio} \geq \rho$, we perform an \text{ID} replacement and update operation. Specifically, all $\text{ID}_{f} \in c_{\mathcal{P}_f}$ present in the $\mathcal{P}_{\text{frame}}$ are replaced with  $\text{ID}_{s} \in \max(\{c_{\mathcal{P}_s}\})$ to get, $\mathcal{P'}_{\text{frame}}$ which is then concatenated to $\mathcal{P}_{\text{scene}}$. Additionally, set of points from $\mathcal{P'}_{\text{scene}}$ can also be deleted to retain constant sparsity, ensuring fixed computation requirement per update. The updated IDs are then appended to the $\mathcal{Q}$; conversely, if the overlap ratio does not meet the threshold requirement, a new entry is added in $\mathcal{Q}$.
\vspace{-0.3cm} 
\begin{figure}[h]
  \centering
    \centering % Centers the image
    \includegraphics[width=0.45\textwidth,keepaspectratio]{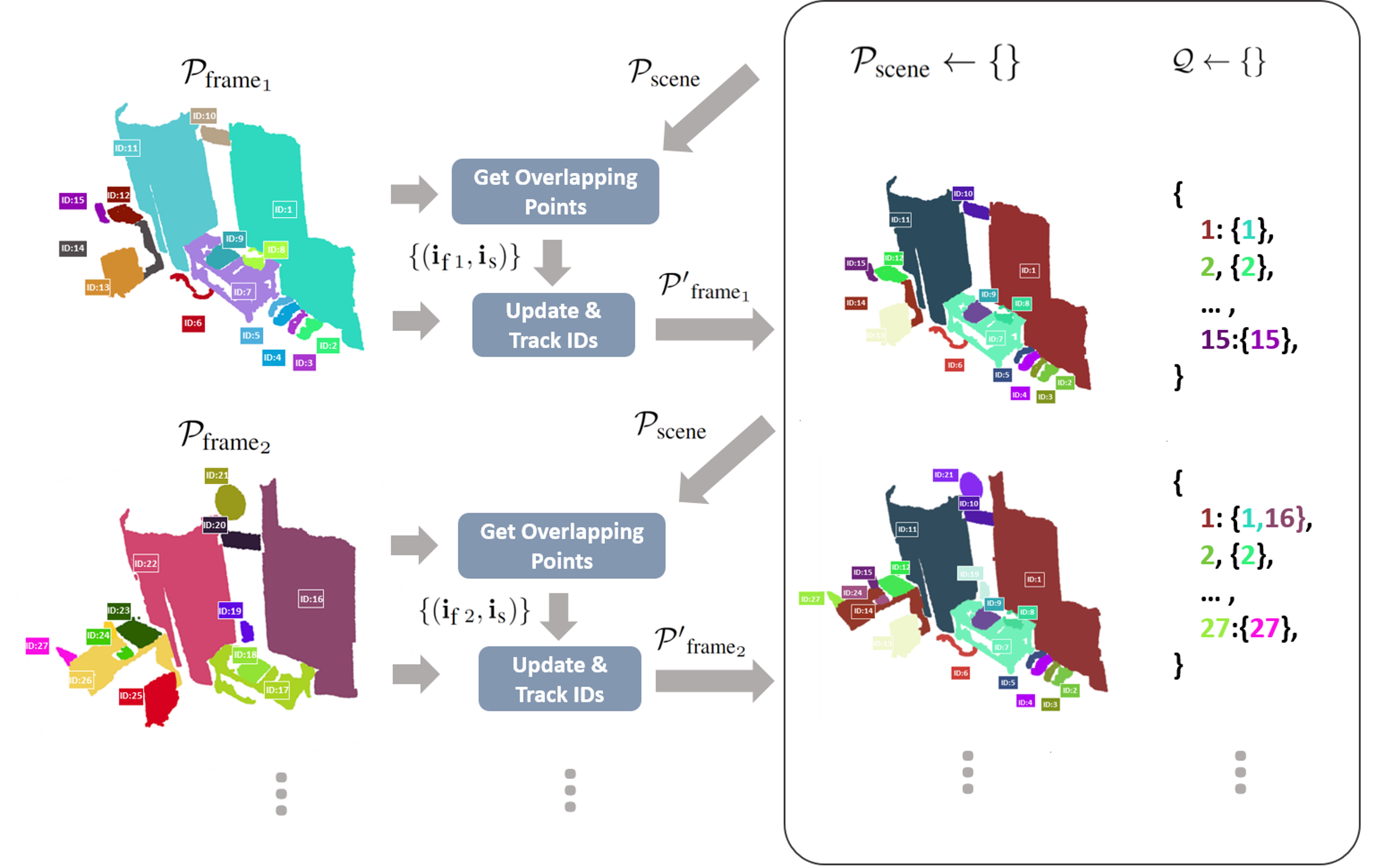}

%%% Explained
%\caption{\textbf{2D-3D Fusion and Tracking:} Tracks and updates IDs of each back-projected 3D semantic mask associated with an image by efficiently assessing the overlap region in 3D space. Tracked IDs are recorded and updated mask-projections concatenated.}

\caption{\textbf{2D-3D Fusion and Tracking.} Tracks IDs of projected 3D semantic masks, associated with instance in image by assessing overlap in 3D space; tracked IDs are recorded and updated mask-projections concatenated.}

%%%\textbf{ Please give more description explaning each step in pipeline that can guide readers clearly}}
%\vspace{-0.5cm} % Adjust this value to reduce the gap

  \label{fig:feat_fuse}
\end{figure}
\vspace{-0.3cm}

\subsubsection{Post Processing}
\label{sec:post_process}

%The point cloud $\mathcal{P}_{\text{scene}}$ with updated IDs, corresponding tracked overlapping IDs $\mathcal{Q}$ along with per-image metadata is processed to construct an instance-centered 3D map 

The point cloud $\mathcal{P}_{\text{scene}}$ with updated IDs, corresponding tracked overlapping IDs $\mathcal{Q}$, along with per-image metadata, is processed into instance-centric map
\newline$\mathcal{M} = \left\{\left(\mathcal{P}, n, c, f_{\text{MV}}, bb_{3D}, (x_{c}, y_{c})\right)_{i} \middle| i \in \texttt{uniq}(\text{ID} \in \mathcal{P}_{\text{scene}})\right\}$.

%\vspace{-0.2cm}
%\begin{equation}
%\begin{aligned}
%\mathcal{M} := & \left\{ \left( \mathcal{P}, n, c, f_{\text{MV}}, bb_{3D}, (x_{c}, y_{c}) \right)_i \mid \right. \\
%& \left. i \in \texttt{uniq}(\text{ID} \in \mathcal{P}_{\text{scene}}) \right\}
%\end{aligned}
%\end{equation}
%\vspace{-0.2cm}

For each distinct 3D object $\mathcal{P}_{i}$, we perform DBSCAN based clustering to reduce noise and achieve fine-grained 3D masks. Accordingly, 3D bounding box $bb_{3D, i}$ and centroid $(x_{c}, y_{c})_{i}$ are computed. For each multiview image corresponding to the 3D object $\mathcal{P}{i}$, names $N'$, captions $C'$, prediction scores $S'_{\text{pred}}$, and feature vectors $F'$ are retrieved using $\mathcal{Q}$ for aggregation and fusion.
The label (name) \( n_i \in N' \) and caption \( c_i \in C' \), corresponding to the highest $S'_{\text{pred}}$ are assigned to 3D instance $\mathcal{P}_{i}$. 
Alternatively, top \( m \) names from \( N' \) based on prediction scores $S'_{\text{pred}}$, are refined using a LLM \cite{gpt4} with prompt: \emph{``assign a single name to the object based on a given list of names''}, yielding more accurate label (name) \( n'_i \). Lastly, feature vector corresponding to multiple views $f_{\text{MV}_{i}}$ is obtained via multiview feature fusion formulation (described in Sec. \ref{sec:feature_fusion}) by top \( m \) fusing feature vectors with based on scores $S'_{\text{pred}}$.

\subsubsection{Feature Fusion}
\label{sec:feature_fusion}
Given a list of feature vectors $\{f\}_{k} $ from multiple scale crops of an instance in an image and feature vectors $\{f_{MS}\}_{m}$ corresponding to multiview image of a 3D instance, a simple and direct fusion scheme aggregates these feature vectors as shown below: 
\vspace{-0.2cm}
\begin{center}
\begin{minipage}{.5\linewidth}
\begin{equation}
f_{\text{MS}} = \frac{1}{k} \sum_{i=1}^{k} f_i 
\label{eq:direct_fusion_multi_scale}
\end{equation}
\end{minipage}%
\begin{minipage}{.5\linewidth}
\begin{equation}
f_{\text{MV}} = \frac{1}{m} \sum_{i=1}^{m} f_{\text{MS}_i} 
\label{eq:direct_fusion_multi_view}
\end{equation}
\end{minipage}
\end{center}
\vspace{-0.2cm}
The fusion schemes in Eq. \ref{eq:direct_fusion_multi_scale} and Eq. \ref{eq:direct_fusion_multi_view}, lack contextual information, resulting in suboptimal performance for contextual/relative queries (Table \ref{tab:sum_fusion_schemes}). \cite{openmask} and our ablations (Sec. \ref{ablation}) indicate that while multiscale crops (important for encompassing contextual information) enhance accuracy, larger crops diminish overall object recall. To mitigate these issues, we propose a modified multiscale feature fusion scheme Eq. \ref{eq:fusion_multi_scale}. This approach employs weighted aggregation of multiscale crop feature vectors, where $f_1$ represents the best-fit crop's feature vector and $\varepsilon$ is a small number $\approx$ 1e-8.
\vspace{-0.2cm}
\begin{equation}
f_{\text{MS}} = \frac{1}{k} \sum_{i=1}^{k} \left( \frac{f_1 \cdot f_i}{\max(\|f_1\|_2 \cdot \|f_i\|_2, \varepsilon) } \right) \cdot f_i \
\label{eq:fusion_multi_scale}
\end{equation}
%\vspace{-0.15cm}
For multiview feature integration, inspired by the per-pixel feature representation in \cite{conceptfusion}, we propose incorporating a global feature vector \( f_\text{G} \) from the entire image while synthesizing our per instance feature representation, defined as:
\vspace{-0.2cm}
\begin{equation}
f_{\text{MV}} = \frac{1}{m} \sum_{i=1}^{m}  f_{\text{MS}_i} + \left( \frac{f_{\text{MS}_i} \cdot f_{\text{G}_i}}{\max(\|f_{\text{MS}_i}\|_2 \cdot \|f_{\text{G}_i}\|_2, \varepsilon)} \right) \cdot f_{\text{G}_i} 
\label{eq:fusion_multi_view}
\end{equation}
\vspace{-0.4cm}

%\subsection{Open Vocabulary Query (Inference)}
\subsection{Instance Retrieval \& Segmentation }
\label{sec:query_method}
Given a map \(\mathcal{M}\), open-vocabulary 3D object search or 3D instance retrieval and segmentation is performed in two stages. First, a query \(\mathcal{K}\) is processed using the CLIP~\cite{clip} text encoder to obtain the feature vector \(f_\mathcal{K}\). Second, cosine similarity scores \(\{S_{\text{score}}\}\) are computed with all 3D instances. The segmentation mask of the 3D instance with the highest similarity score is retrieved as the most likely response to \(\mathcal{K}\).

\subsection{Spatial Reasoning }
\label{sec:method_sr}
For queries requiring complex spatial reasoning, our approach involves in-context learning, leveraging large context window of LLM \cite{gpt4},
enabling complex spatial reasoning based on coherent 3D representations and associated metadata such as mask labels, centroids, bounding boxes and detailed caption available for the constructed scene $\mathcal{M}$.
%to perform spatial reasoning based on coherent 3D representation and metadata such as mask labels, centroids, bounding boxes and description available for the constructed scene $\mathcal{M}$.
%Our approach enables LLM to perform complex spatial reasoning, by providing reasoning rules, and coherent 3D representation (centroids, bounding boxes, labels, detailed captions) as context to LLM.
%By providing the LLM with a coherent 3D representation (labels, centroids, bounding boxes, captions), and reasoning rules, our approach enables LLM to perform complex spatial reasoning.
$\mathcal{M}' := \mathcal{M} \setminus \{\mathcal{P}, f_{\text{MV}}\}$ is provided to LLM along with a system prompt, which is crafted using the following prompting strategy:

%For queries involving complex spatial reasoning, the approach leverages the long context length of Large Language Models like GPT-4V \cite{gpt4}. A strategy is employed where a simplified map \(\mathcal{M}' := \mathcal{M} \setminus \{\mathcal{P},f,f^g\}\) is parsed to the LLM, using the prompting strategy defined as:

\begin{itemize}
    \item \text{Use `Name' \& `Description' to understand object.}
    \item \text{Use `ID' to refer object.} 
    \item \text{Use `Cartesian Coordinates'.} 
    \item \text{Get `Centroid' \& `Bounding Box' information.} 
    \item \text{Compute `Euclidean Distance' if necessary.} 
    \item \text{Assume `Tolerance' if necessary.} 
\end{itemize}

%\section{EXPERIMENTS}
%\section{RESULTS AND DISCUSSION}
%\label{ch:results}

% target -> 85-100 lines
\section{Experiments}

\subsection{Implementation Details}

\subsubsection{Models Utilized}

GroundedSAM \cite{gsam} (a method based on RAM++ \cite{ram++} (ram\_plus\_swin\_large\_14m), GroundingDINO \cite{gdino} (groundingdino\_swint\_ogc), and SAM \cite{sam} (sam\_vit\_h\_4b8939)) is employed for generating instance segmentation masks and bounding boxes. GPT-4V \cite{gpt4} (gpt-4-vision-preview, gpt-4-1106-preview) is used for detailed captions \& names and spatial reasoning. The CLIP encoder \cite{clip} (ViT-H-14 pre-trained on laion2b\_s32b\_b79k dataset) is used for instance feature vectors. % For spatial reasoning and annotation tasks, we leverage the powerful GPT-4 \cite{gpt4} language  model.

%a method based on RAM++ \cite{ram++} (ram\_plus\_swin\_large\_14m), GroundingDINO \cite{gdino} (groundingdino\_swint\_ogc), and SAM \cite{sam} (sam\_vit\_h\_4b8939)

\subsubsection{Hyper parameter Settings}
Hyper-parameters, determined through ablation studies on Replica \cite{replica} (Sec: \ref{ablation}), are consistent across datasets. We set \( m=5 \) to select the top 5 images and use \( k=3 \) with 3 crop levels and a scaling ratio increment of 0.2 (\( S_{r} = [0.8, 1, 1.2] \)). A stride of \( s=40 \) ensures frame overlap. For GroundedSAM \cite{gsam}, thresholds are: IoU-0.4, bounding box-0.25, and text-0.25. Padding of \( px=20 \) pixels delineates instance mask borders. Overlap evaluation uses voxel size \( \epsilon=0.02 \) and threshold \( \rho=0.3 \). DBSCAN uses epsilon 0.1 and a minimum cluster size of 20 points. GPT-4 \cite{gpt4} is set with a temperature of 0.

\subsubsection{Filtering and Post-Processing}
\label{sec:filterpp}

Large background objects (walls, ground, roof, ceiling) and objects with bounding boxes $>$ 95\% of image area are excluded to prevent their feature vectors from exhibiting similarity to foreground objects, adversely affecting recall and score distribution. In DBSCAN post-processing, clusters with points $\geq$ 80\% of the largest cluster are treated as separate instances with unique IDs and attributes. In cases where an object is undetectable by GPT-4 \cite{gpt4}, instances are assigned RAM++ \cite{ram++} names and given simplified captions: \emph{``an \{object\} in a scene''}.

 %During GPT-4 \cite{gpt4} safety mode, instances use RAM++ \cite{ram++} names and simplified captions: ``an {object} in a scene".

\subsection{Evaluation Details}

\subsubsection{Datasets}

Multiple scenes from the semi-synthetic dataset Replica~\cite{replica} (\emph{room0}, \emph{room1}, \emph{room2}, \emph{office0}, \emph{office1}, \emph{office2}, \emph{office3}, \emph{office4}) and real-world dataset ScanNet~\cite{scannet} (\emph{scene0000\_00}, \emph{scene0034\_00}, \emph{scene0164\_03}, \emph{scene0525\_01}, \emph{scene0549\_00}) were used for thorough qualitative and quantitative evaluation. Similar to previous studies~\cite{conceptgraph, conceptfusion}, a limited number of scenes were selected due to the extensive manual human evaluation.

%Similar to other similar work ~\cite{conceptgraph, conceptfusion} limited number of scenes were selected due extensive manual human evaluation.

\subsubsection{Quantitative Evaluation}
\label{sec:exp_quatnt}

The proposed method is evaluated using standard metrics: mean recall accuracy (mAcc), frequency-weighted IoU (F-mIoU), and average precision (AP) at IoU thresholds [0.5:0.05:0.95], including AP50 and AP25 from ScanNet \cite{scannet}. For open vocabulary performance, as in \cite{openmask, conceptgraph}, 3D masks were retrieved with ground truth labels and the prompt: \emph{``an \{object\} in a scene''}. Masks were downsampled to 0.25cm voxel size, followed by nearest neighbor search for intersecting points. Results on the Replica \cite{replica} dataset were compared against state-of-the-art models \cite{openmask, segment3d, conceptfusion, conceptgraph}, using identical prompts and foundation models and simple feature fusion formulation (Eq. \ref{eq:direct_fusion_multi_view} and Eq. \ref{eq:direct_fusion_multi_scale}).

\subsubsection{Qualitative Evaluation}
\label{sec:experiment_setting_qualitative_evaluation}
Extensive qualitative assessments for open vocabulary instance retrieval, annotation, segmentation, and spatial reasoning were performed through manual human evaluation. For \textbf{open vocabulary instance retrieval}, over 1,000 queries regarding instances (e.g. \emph{``musical instrument''}), affordances (e.g., \emph{``place to sit and work''}), properties (e.g., \emph{``green towel''}), and relative queries (e.g., \emph{``green towel next to sink''}) were made. 
Performance was evaluated based CLIP~\cite{clip} based instance retrieval (see Sec. ~\ref{sec:query_method}) for four feature fusion schemes: \emph{Scheme 1} represents direct aggregation of multiscale (Eq. \ref{eq:direct_fusion_multi_scale}) and multiview features (Eq. \ref{eq:direct_fusion_multi_view}), \emph{Scheme 2} represents updated multiview features (Eq. \ref{eq:fusion_multi_view}), \emph{Scheme 3} represents updated multiscale features (Eq. \ref{eq:fusion_multi_scale}) with increased crop expansion ratios (\(S_r = [1, 2, 4]\)), and \emph{Scheme 4} represents combination of updated multiview (Eq. \ref{eq:fusion_multi_view}) and multiscale (Eq. \ref{eq:fusion_multi_scale}) features fusion formulation. The \textbf{annotation and segmentation} capabilities of the proposed approach were evaluated through manual verification of label assignments and mask merging. For \textbf{spatial reasoning}, 70 complex reasoning questions (e.g., \emph{``I'm feeling cold what should I do?''}) were administered across all scenes using a large language model (see Sec. ~\ref{sec:method_sr}), assessing viability  of proposed prompting strategy for spatial reasoning.

\vspace{-0.2cm}

\subsection{Results and Discussion}
\label{sec:result_and_analysis}

\subsubsection{Ablation Studies}
\label{ablation}
% should be in appendix

To assess the influence of hyperparameters, using \cite{openmask} experiment setup; ablation studies were conducted on Crop Level $k$, Top Images $m$, and Crop Ratios $S_{r}$ using quantitative metrics. Top Images $m$ affects multiview feature fusion (Eq. \ref{eq:fusion_multi_view}), representing the feature vectors for aggregation. Crop Ratio $S_{r}$ and Crop Level $k$ affect multiscale feature fusion (Eq. \ref{eq:direct_fusion_multi_scale}), determining crop size and quantity for feature vector aggregation. Crop Level $k$ amplifies Crop Ratio $S_{r}$'s effect, as a higher $k$ with the same $S_{r}$ results in larger crops.

Similar to \cite{openmask}, we found that extreme values of these hyperparameters deteriorate results. A low $m$ reduces redundancy, while a high $m$ may include bad images, as shown in Table \ref{tab:ablation}. Lower values of $S_r$ and $k$ may not harm the model but could introduce redundancy. Larger $S_r$ values add context, but can saturate the distribution of similarity scores. 

%We select $m=5$, $k=3$, and crop levels with a $0.2$ increment in scaling ratio i.e $S_{r} = [0.8, 1, 1.2]$.
%\textbf{Lastly add a concluding statement for each experiment like, based on the table xx, m = 5, sr = xx offers best performance compared to others...for all experiments if not added}

%\textbf{Also refer to the website and talk about various results you presented on the website and give the link again in experiments}

\begin{table}[H]
\setlength{\tabcolsep}{2pt}
\renewcommand{\arraystretch}{0.85}
\centering
\begin{tabular}{@{}ccccccc@{}}
\toprule
\multirow{2}{*}{\text{Parameter}} & \multirow{2}{*}{\text{Value}} & \multicolumn{5}{c}{\text{Replica \cite{replica}}} \\
\cmidrule(lr){3-7}
 & & mAcc & F-mIoU & AP & AP50 & AP25 \\
\midrule
\multirow{3}{*}{Top Images ($m$)} & 1.0 & 39.6 & 43.4 & 8.7 & 19.3 & 27.2 \\
 & \textbf{5.0} & \textbf{40.8} & \textbf{44.7} & \textbf{8.9} & \textbf{19.6} & \textbf{27.7} \\
 & 10.0 & 39.3 & 44.3 & 8.7 & 19.1 & 27.5 \\
\midrule
\multirow{3}{*}{Crop Levels ($k$)} & 1.0 & 35.9 & 43.6 & \textbf{9.1} & \textbf{19.6} & \textbf{27.7} \\
 & \textbf{3.0} & \textbf{40.8} & \textbf{44.7} & 8.9 & \textbf{19.6} & \textbf{27.7} \\
 & 5.0 & 39.4 & 44.3 & 8.8 & 19.4 & 26.9 \\
\midrule
\multirow{3}{*}{Crop Ratio ($S_{r}$)} & [0.1,1,1.1] & 39.9 & 44.4 & \textbf{8.9} & 19.4 & \textbf{28.1} \\
 & \textbf{[0.8,1,1.2]} & \textbf{40.8} & \textbf{44.7} & \textbf{8.9} & \textbf{19.6} & 27.7 \\
 & [0.7,1,1.3] & 39.9 & 44.8 & \textbf{8.9} & 19.4 & 27.3 \\
\bottomrule
\end{tabular}%

\caption{\textbf{Ablation study of Hyperparameters.} The total images $m$ w.r.t best prediction scores ($s_{pred}$), for multiview feature fusion. Increment in ratio $S_{r}$ for scaling crop sides and total number of crops $k$, for multiscale feature fusion.}
%\caption{\textbf{Ablation study of Hyperparameters.} The total images $m$ w.r.t best $s_{pred}$(\textbf{this is not present in context}), for multiview feature fusion, Ratio $S_{r}$ \textbf{(use crop ratio wording)} for scaling crop sides and total number of crops $k$(\textbf{and crop scale wording to be consistent}), considered for multiscale feature fusion.}
\label{tab:ablation}
\end{table}

%Additionally, an ablation study with different CLIP model variants was also conducted. As shown in Table \ref{tab:clip_models}, larger CLIP variants improved overall performance. As an example, Fig. \ref{fig:clip_compare} illustrates that a larger CLIP model better associates properties with the queried object, distinguishing between \emph{``single sofa''} and \emph{``double sofa''}.

%\input{content/Tables/ablation_studies/sum/clip_models}

%\input{content/images/appendix/model}

%\subsection{Open Vocabulary Instance Retrieval}

\subsubsection{Quantitative Comparison with Baseline Methods}

The proposed method demonstrates comparable or better performance than baselines on quantitative metrics, as shown in Tables~\ref{tab:cg_replica} and~\ref{tab:o3d_replica}. These tables compare segmentation mask accuracy and precision in response to open vocabulary queries against ground truth masks. %We used Eq. \ref{eq:direct_fusion_multi_scale} and \ref{eq:direct_fusion_multi_view} for direct feature aggregation and
We followed the original ConceptGraph~\cite{conceptgraph} and OpenMask3D~\cite{openmask} setups for fair assessment. Overall, our method performed on par or better across all metrics and datasets.

%Under conditions similar to those outlined in baseline studies, the proposed method demonstrates comparable or better performance on quantitative metrics. We report results in Tables \ref{tab:cg_replica} and \ref{tab:o3d_replica}, comparing them to those reported in the respective original studies. The objective is to assess the accuracy and precision in segmentation masks retrieved in response to open vocabulary queries, with respect to the ground truth masks.

%For comparisons with methods listed in Table \ref{tab:cg_replica}, we employ \texttt{CLIP-ViT-H} as the CLIP encoder and invoke instances with the prompt \texttt{``an image of a \{object\}''}, as described in ConceptGraph \cite{conceptgraph}. For the methods outlined in Table \ref{tab:o3d_replica}, we utilize the prompt \texttt{``an \{object\} in scene''} with \texttt{CLIP-ViT-L} as the CLIP encoder, in accordance with the original study OpenMask3D \cite{openmask}, further details regarding experimental setup are discussed in Sec. \ref{sec:exp_quatnt}. Overall, across all metrics and datasets, we found the performance to be comparable to or better than the baseline work.

\begin{table}[h]
\setlength{\tabcolsep}{2pt}
\renewcommand{\arraystretch}{0.9}
\centering

% Table generated by Excel2LaTeX from sheet 'baseline comp (2)'
\begin{tabular}{lcccc}
\toprule
\multicolumn{2}{c}{\multirow{2}[4]{*}{Method}} &   & \multicolumn{2}{c}{Replica \cite{replica}} \\
\cmidrule{4-5}\multicolumn{2}{c}{} &   & mAcc & F-mIoU \\
\midrule
ConceptFusion \cite{conceptfusion} &   &   & 24.2 & 31.3 \\
ConceptFsuion+SAM \cite{conceptfusion} &   &   & 31.5 & 38.7 \\
ConceptGraph \cite{conceptgraph} &   &   & 40.6 & 36.0 \\
ConceptGraph-Detector \cite{conceptgraph} &   &   & 38.7 & 35.4 \\
OpenSU3D (Ours) &   &   & \textbf{42.6} & \textbf{40.9} \\
\bottomrule
\end{tabular}%

\caption{\textbf{Comparison of open-vocabulary segmentation results} with ConceptGraph\cite{conceptgraph} setup.}

\vspace{-0.2cm}
%\vspace{-0.5cm} % Adjust this value to reduce the gaps
\label{tab:cg_replica}
\end{table}
\begin{table}[h]
\setlength{\tabcolsep}{2pt}
\renewcommand{\arraystretch}{0.9}
\centering

\begin{tabular}{lcccc}
\toprule
\multicolumn{1}{c}{\multirow{2}[4]{*}{Method}} &   & \multicolumn{3}{c}{Replica \cite{replica}} \\
\cmidrule{3-5}  &   & AP & AP50 & AP25 \\
\midrule
OpenMask3D \cite{openmask} &   & \textbf{13.0} & 18.4 & 24.2 \\
OpenMask3D+Segment3D \cite{segment3d} &   & - & 18.7 & - \\
OpenSU3D (Ours) &   & 8.9 & \textbf{19.6} & \textbf{27.7} \\
\bottomrule
\end{tabular}%

%\caption{\textbf{Comparison of open-vocabulary segmentation  results}. Quantitative evaluation of open vocabulary instance segmentation with counterpart methods using the experimental setup from OpenMask3D \cite{openmask}. }

\caption{\textbf{Comparison of open-vocabulary segmentation results} with OpenMask3D \cite{openmask} setup. }

\vspace{-0.2cm}

%\vspace{-0.5cm} % Adjust this value to reduce the gap

\label{tab:o3d_replica}
\end{table}

\vspace{-0.2cm}
\subsubsection{Qualitative Comparison with Baseline Methods}

The quantitative evaluation was primarily designed for closed vocabulary assessments, relying on recall accuracy that does not reflect real-world requirements for open vocabulary queries. Additionally, these evaluations, dependent on the quantity of mask proposals \cite{segment3d}, might not accurately represent true performance against closed set ground truth masks.

\begin{figure}[h]
  \centering

   \cg{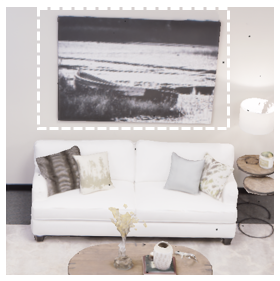}{a picture on wall}
   \od{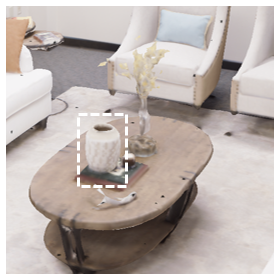}{an empty vase}
   %\cg{2}{a coffee table} 
   %\od{11}{fridge}

  %\cg{3}{round table}
  
  %\od{2}{an empty vase}

  %\od{11}{fridge}

    %\caption{\textbf{Heatmaps representing similarity between text queries and scene-instances.} For a given text query (left), comparison of per instance cosine similarity scores for OpenSU3D (middle) \& baseline methods (left). `Dark red' represents maximum similarity, and `dark blue' indicates minimum similarity.} 
\caption{\textbf{Text-Instance Similarity Heatmaps.} Cosine similarity for text queries using ConceptGraph \cite{conceptgraph}, OpenMask3D \cite{openins3d} and our method. \textcolor{red!50!black}{\ensuremath{\blacksquare}}: max, \textcolor{blue!20!black}{\ensuremath{\blacksquare}}: min similarity.}

%\vspace{-0.5cm} % Adjust this value to reduce the gap

\vspace{-0.6cm}    

  \label{fig:baseline_compare}
\end{figure}

To address these limitations, we provide comprehensive qualitative comparisons with baseline works in Fig. \ref{fig:baseline_compare}. The goal is to assess the ability to recall the correct segmentation mask for open vocabulary queries, assigning high similarity scores to relevant objects and lower scores to irrelevant ones.
%\textbf{You can also add that there are many qualitative experiments reported on the website}
Specifically, our proposed method showed better 2D to 3D association and distribution of similarity scores with proposed multiscale and multiview feature fusion formulation (Eq. \ref{eq:fusion_multi_scale} and Eq. \ref{eq:fusion_multi_view}). As shown in Fig. \ref{fig:baseline_compare}, for the queries \emph{``a picture on wall''} and \emph{``an empty vase''}, both baseline methods recalled incorrect objects, while our method works perfectly.
%In contrast, for the queries \emph{``a coffee table''} and \emph{``fridge''}, the proposed approach demonstrated better similarity score distribution, assigning higher similarity to relevant objects.
%Additional results: \href{https://opensu3d.github.io/#comparision}{\textit{opensu3d.github.io/#comparision}}

\vspace{-0.2cm}
\subsubsection{Assessment of Feature Fusion Schemes}
We conducted a qualitative evaluation of feature fusion schemes as defined in Sec. \ref{sec:experiment_setting_qualitative_evaluation}. The results are summarized in Table \ref{tab:sum_fusion_schemes}.

\begin{figure}[H]
  \centering
    %\schm{5}{pan on stove}

    \schmshort{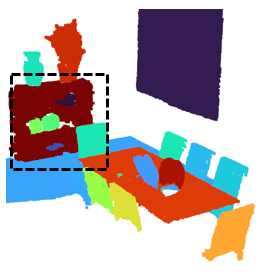}{plant on shelf}

    \schmshort{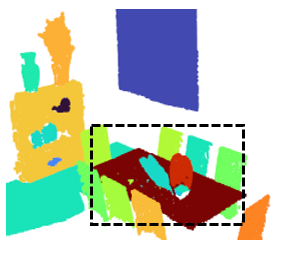}{plant on table}

    %\schm{5}{pan on stove}

    %\schm{8}{schoolbag on bed}

\caption{\textbf{Relative Query Similarity Heatmaps.} For a given text query, per-instance cosine similarity heatmaps across feature fusion schemes. \textcolor{red!50!black}{\ensuremath{\blacksquare}}: max, \textcolor{blue!20!black}{\ensuremath{\blacksquare}}: min similarity.}

\vspace{-0.5cm}

  \label{fig:method_schemes}
\end{figure}

For {instance}, {property}, and {affordance} queries, performance across all schemes was similar. However, for {relative} queries, \emph{Scheme 2} and \emph{Scheme 3} with our proposed multiscale and multiview fusion formulations (Eqs. \ref{eq:fusion_multi_view} and \ref{eq:fusion_multi_scale}) outperformed \emph{Scheme 1}. \emph{Scheme 4}, incorporating both of the proposed formulations, achieved the best recall accuracy. Additionally, similarity score heatmaps as shown in Fig. \ref{fig:method_schemes} shows that \emph{Scheme 1} often wrongly assigned the highest score to the largest instance. In contrast, the updated fusion formulations in \emph{Schemes 2, 3,} and \emph{4} improved the recall of instance masks and similarity score distribution, with \emph{Scheme 4} performing the best overall.

\begin{table}[h]
\setlength{\tabcolsep}{2pt}
\renewcommand{\arraystretch}{0.9}
\centering

% Table generated by Excel2LaTeX from sheet 'qualitative-manual (2)'
\begin{tabular}{ccccccrcccc}
\toprule
\multicolumn{1}{c}{\multirow{2}[4]{*}{Feature Fusion \newline{}}} &   & \multicolumn{4}{c}{Replica \cite{replica}} &   & \multicolumn{4}{c}{ScanNet \cite{scannet}} \\
\cmidrule{3-6}\cmidrule{8-11}  &   & Inst. & Aff. & Prop. & Rel. &   & Inst. & Aff. & Prop. & Rel. \\
\midrule
Scheme 1 &   & 0.8 & 0.7 & 0.7 & 0.3 &   & 0.8 & \textbf{0.8} & 0.7 & 0.4 \\
Scheme 2 &   & 0.8 & 0.7 & \textbf{0.9} & 0.5 &   & \textbf{0.9} & 0.7 & \textbf{0.8} & 0.6 \\
Scheme 3 &   & \textbf{0.9} & \textbf{0.9} & \textbf{0.9} & \textbf{0.6} &   & \textbf{0.9} & \textbf{0.8} & 0.7 & 0.6 \\
Scheme 4 &   & 0.8 & \textbf{0.9} & \textbf{0.9} & \textbf{0.6} &   & \textbf{0.9} & 0.7 & 0.7 & \textbf{0.7} \\
\bottomrule
\end{tabular}%

\caption{\textbf{Evaluation of feature fusion schemes}. Accuracy of fusion schemes for retrieval with ``Inst.'' (instance), ``Aff.'' (affordance), ``Prop.'' (property), and ``Rel.'' (relative) text queries, as assessed by a human evaluator.}

%\caption{\textbf{Qualitative evaluation of feature fusion schemes on open vocabulary instance retrieval performance}. Accuracy of feature fusion schemes for instance retrieval with ``Inst.'' (instance), ``Aff.'' (affordance), ``Prop.'' (Property). and ``Rel.'' (relative) text queries, as assessed by a human evaluator. }

\vspace{-0.5cm}

\label{tab:sum_fusion_schemes}
\end{table}

\vspace{-0.2cm}

%1. Human Eval, Qualitative Results

%\input{content/images/results/queries}

\vspace{-0.2cm}
\subsubsection{Open Set Annotation and Segmentation}
\label{sec:exp_os}

Annotation accuracies for directly assigned labels $n$ using the maximum prediction score $S'_{\text{pred}}$ and labels $n'$, where top $m$ labels based on $S'_{\text{pred}}$ are refined by LLM (see Sec. \ref{sec:post_process}), were manually verified across all Replica \cite{replica} and ScanNet \cite{scannet} scenes. 
To evaluate open set segmentation, mask merging accuracy is determined by counting and classifying under-merges and over-merges as faulty. Evaluations summarized in Table \ref{tab:sum_label_accuracy} show that LLM labels $n'$ are more accurate than direct labels $n$. Additionally, this helps to filter out undesired large instances (see Sec. \ref{sec:filterpp}), resulting in slight improvement in mask merging accuracy. Furthermore, we found that LLM labels $n'$ are more concise than direct labels $n$.

%Annotation accuracy for directly assigned label $n$ using maximum prediction score  $S'_{\text{pred}}$  and label $n^\prime$, LLM refined top $m$ labels based on $S'_{\text{pred}}$ (discussed in \ref{sec:post_process}); were manually verified across all Replica \cite{replica} and ScanNet \cite{scannet} scenes. To evaluate open set segmentation, under-merges and over-merges were counted and classified as faulty.
%Evaluations summarized in Table \ref{tab:sum_label_accuracy} shows that LLM labels $n^\prime$ are more accurate than direct labels $n$, highlighting that alternate approach with LLM for label refinement improves label (name) assignment while filtering undesired large instances (Sec. \label{sec:filterpp}). This leads to a slight improvement in mask merging accuracy. Additionally, we found LLM labels $n^\prime$ are more concise than direct labels $n$. 

\begin{table}[h]
\setlength{\tabcolsep}{2pt}
\renewcommand{\arraystretch}{0.9}
\centering

% Table generated by Excel2LaTeX from sheet 'qualitative-manual (2)'
\begin{tabular}{ccccccccc}
\toprule
\multicolumn{1}{c}{\multirow{3}{*}{Labels}} &   & \multicolumn{2}{c}{Replica \cite{replica}} &   & \multicolumn{2}{c}{ScanNet \cite{scannet}} \\
\cmidrule{3-4}\cmidrule{6-7}  &   & \multicolumn{1}{c}{\multirow{2}{*}{Label Acc.}} & \multicolumn{1}{c}{\multirow{2}{*}{Merge Acc.}} &   & \multicolumn{1}{c}{\multirow{2}{*}{Label Acc.}} & \multicolumn{1}{c}{\multirow{2}{*}{Merge Acc.}} \\
  &   &   &   &   &   &  \\
\midrule
Direct Label ($n$)  &   & 0.83 & 0.87 &   & 0.75 & 0.85 \\[0.1cm]
LLM Label ($n'$)  &   & \textbf{0.87} & \textbf{0.88} &   & \textbf{0.84} & \textbf{0.87} \\
\bottomrule
\end{tabular}

\caption{\textbf{Qualitative evaluation of segmentation and annotation accuracy}. For Direct Label ($n$) and LLM Label ($n'$), the annotation and merge accuracy of segmentation masks, as assessed by a human evaluator.}

\vspace{-0.2cm}

\label{tab:sum_label_accuracy}
\end{table}

\vspace{-0.15cm}

%1.assess by human evaluator
\subsubsection{Complex Spatial Reasoning}

To assess spatial reasoning, as specified in Sec.~\ref{sec:experiment_setting_qualitative_evaluation}, we posed complex spatial reasoning questions (Example,  Fig.~\ref{fig:flow} \emph{``Which sink is closer to bed''}) across all scenes.
%To assess spatial reasoning, we posed 70 complex spatial questions (Example,  Fig.~\ref{fig:flow} ``Which sink is closer to bed'') across scenes as specified in Sec.~\ref{sec:method_sr}. 
The manual assessment showed that, with our approach (see \ref{sec:method_sr}), the LLM~\cite{gpt4} demonstrated effective reasoning ability in 3D space over constructed representation. LLM exhibited higher accuracy in scenes from Replica \cite{replica} (0.83), compared to ScanNet \cite{scannet} (0.68). 
This decline in performance can be attributed to a comparatively higher incidence of flaws in merging and label assignment in larger ScanNet \cite{scannet} scenes.
%This decline in performance in larger scenes can be attributed to a higher incidence of merging flaws and labels association to larger scenes.

\vspace{-0.1cm}

\subsection{Limitations}
\label{sec:limitations}

The effectiveness of this approach is constrained by the capabilities of its underlying foundation models and the occurrence of merging errors. Utilizing CLIP~\cite{clip} for image-text association and ~\cite{gsam} for 2D mask generation, performance correlates with these model's robustness. Moreover, annotation accuracy and spatial reasoning depends on LLM's \cite{gpt4} ability and context length, and are susceptible to occasional merging faults.
%impacted by occasional merging faults.
%are susceptible to occasional merging faults.

\vspace{-0.3cm}

%, which can arise from diverse scene conditions. 

%\subsection{Future Research}

%Higher-order representations like neural \cite{langsplat, lerf} or voxel-based \cite{kimera, voxblox} can enhance segmentation and applications. Open world 3D understanding is nascent, with this study addressing foundational questions. Future work includes applying findings to outdoor scenes and large-scale modeling.

%Using 3D asset generation models like Point-to-3D \cite{pointto3d}, the proposed method could enable applications like 3D scan to digital twin. Integrating this method into SLAM could create temporal records of instances. Coupled with large context window models like Gemini 1.5 \cite{gemani15}, it could enhance 3D spatio-temporal reasoning \cite{lifelongmemory} for dynamic 4D world understanding.

\vspace{-0.05cm}
\section{CONCLUSION}
\vspace{-0.1cm}
In conclusion, this study presents a scalable and incremental framework for constructing open set 3D scene representation for open world 3D scene understanding tasks, addressing the limitations of current methods. By leveraging 2D foundation models, our approach constructs detailed instance-level 3D scene representations, efficiently tracking and associating instance-specific information such as feature vectors, names, and captions. The proposed feature fusion schemes encompass contextual information, enhancing performance on relative queries. Additionally, the use of large language models facilitates robust automatic annotation and enables complex spatial reasoning over 3D scenes. Comprehensive evaluations show that our method achieves superior zero-shot generalization compared to state-of-the-art solutions. 
In the future, we plan to explore spatio-temporal reasoning in 3D dynamic scenes and extend the method from indoors to large-scale outdoor environments.

\vspace{-0.1cm}


%%%%%%%%%%%%%%%%%%%%%%%%%%%%%%%%%%%%%%%%%%%%%%%%%%%%%%%%%%%%%%%%%%%%%%%%%%%%%%%%

%%%%%%%%%%%%%%%%%%%%%%%%%%%%%%%%%%%%%%%%%%%%%%%%%%%%%%%%%%%%%%%%%%%%%%%%%%%%%%%%

%%%%%%%%%%%%%%%%%%%%%%%%%%%%%%%%%%%%%%%%%%%%%%%%%%%%%%%%%%%%%%%%%%%%%%%%%%%%%%%%
%\section*{APPENDIX}


%\section*{ACKNOWLEDGMENT}

%The first author gratefully acknowledge the invaluable guidance of our advisors and the computational resources provided by Huawei Technologies, as well as the support from the Technical University of Munich (TUM). %Their contributions and resources have been instrumental in the successful completion of this research.

%%%%%%%%%%%%%%%%%%%%%%%%%%%%%%%%%%%%%%%%%%%%%%%%%%%%%%%%%%%%%%%%%%%%%%%%%%%%%%%%

\bibliography{references}

\end{document}